\documentclass[10pt,twocolumn,letterpaper]{article}

\usepackage{iccv}
\usepackage{times}
\usepackage{epsfig}
\usepackage{graphicx}
\usepackage{amsmath}
\usepackage{amssymb}
\usepackage{paralist}
\usepackage{color}
\newcommand{\mydataset}{\textit{Auditory Vehicle Tracking}\xspace}

\newcommand{\myapproach}{cross-modal auditory localization\xspace}

\usepackage[pagebackref=true,breaklinks=true,letterpaper=true,colorlinks,bookmarks=false]{hyperref}

\iccvfinalcopy 

\def\iccvPaperID{1083} 

\ificcvfinal\fi
\begin{document}

\title{Self-supervised Moving Vehicle Tracking with Stereo Sound}
\author{Chuang Gan$^{2,3}$, Hang Zhao$^{1}$, Peihao Chen$^{3}$, David Cox$^{2,3}$, Antonio Torralba$^{1}$\\
$^1$ MIT CSAIL,
$^2$ MIT-IBM Watson AI Lab,
$^3$ IBM Research AI}

\maketitle

\begin{abstract}

Humans are able to localize objects in the environment using both visual and auditory cues, integrating information from multiple modalities into a common reference frame.  We introduce a system that can leverage unlabeled audiovisual data to learn to localize objects (moving vehicles) in a visual reference frame, purely using stereo sound at inference time.  Since it is labor-intensive to manually annotate the correspondences between audio and object bounding boxes, we achieve this goal by using the co-occurrence of visual and audio streams in unlabeled videos as a form of self-supervision, without resorting to the collection of ground truth annotations. In particular, we propose a framework that consists of a vision ``teacher'' network and a stereo-sound ``student'' network. During training, knowledge embodied in a well-established visual vehicle detection model is transferred to the audio domain using unlabeled videos as a bridge. At test time, the stereo-sound student network can work independently to perform object localization using just stereo audio and camera meta-data, without any visual input. Experimental results on a newly collected \mydataset dataset verify that our proposed approach outperforms several baseline approaches. We also demonstrate that our \myapproach approach can assist in the visual localization of moving vehicles under poor lighting conditions.

\end{abstract}

\section{Introduction}

Sound conveys a wealth of information about the physical world around us, and humans are remarkably good at interpreting sounds produced by nearby objects. We can often identify what an object is based on the sounds it makes (\eg a dog barking), and we can estimate properties of materials (\eg if they are hard or soft) based on the sounds they make when they interact with other objects.

In addition, our perception of sound allows us to localize objects that are not in our line of sight (\eg objects that are behind us, or that are occluded), and sound plays an important role in allowing us to localize objects in poor lighting conditions. Importantly, our senses of sight and hearing are fundamentally integrated and co-registered---for instance, we can localize an object and accurately point to it, whether we see it, or hear it with our eyes closed. This registration of auditory and visual information into a common reference frame gives us the ability to integrate audio and visual information together when both are present, or to rely on just one when the other is absent.

\begin{figure}[t]
   \centering
   \includegraphics[width = 1\linewidth]{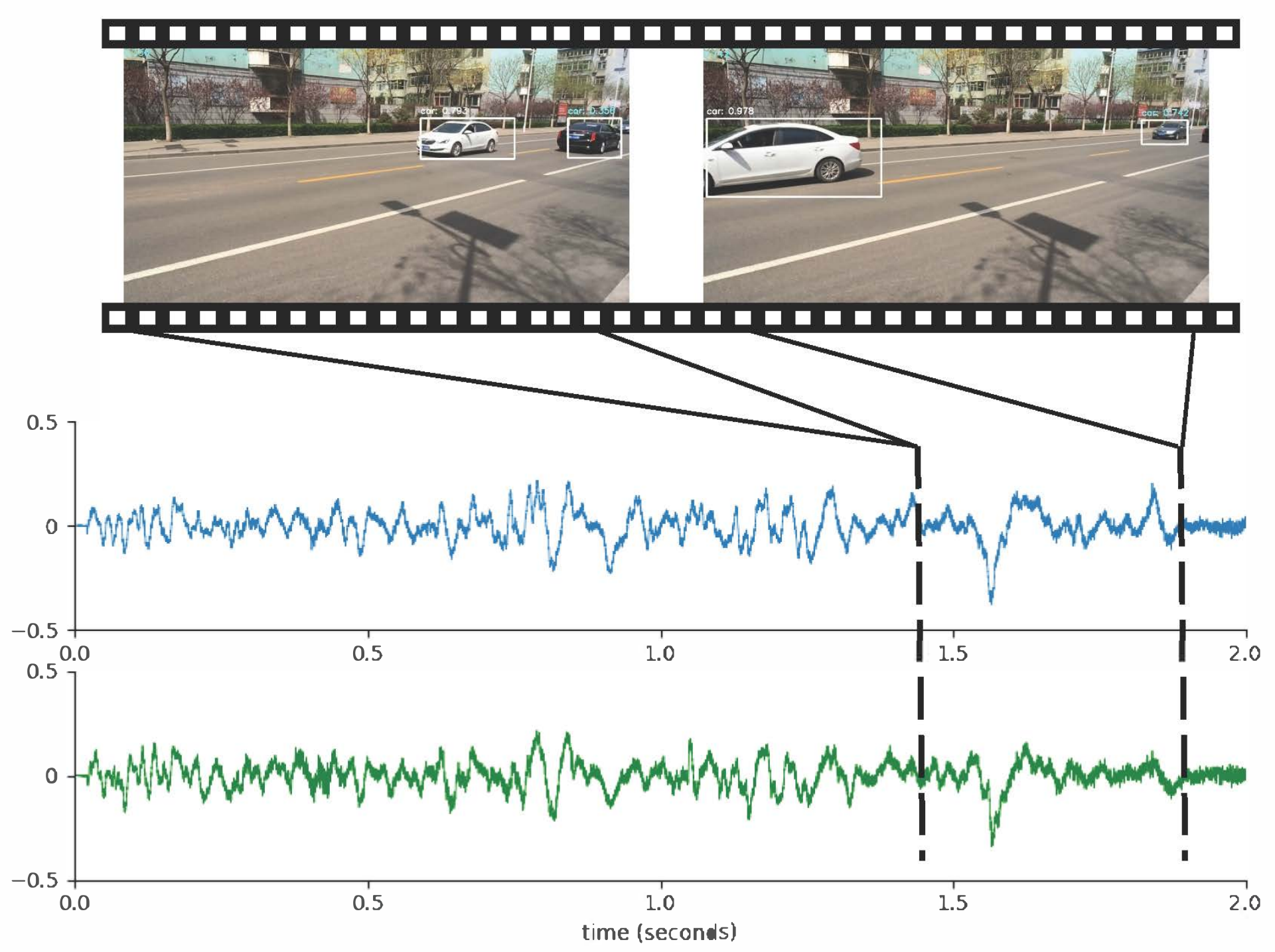}

   \caption{Taking the stereo sounds as input, our proposed \myapproach system can recover the coordinates of moving vehicles in the reference frame purely from stereo sound and camera meta-data, without any visual input.}
   \label{fig:teaser}
\end{figure}

Here, we seek to build a system that can learn auditory-visual correspondences in a self-supervised way, allowing us to perform a classic visual object detection task---drawing bounding boxes around target vehicles---using audio and camera meta-data information alone. Stereo audio provides rich information about the location of objects, due to arrival time and sound level differences between two spatially separated microphones. Figure~\ref{fig:teaser} gives an example to illustrate the setting of the problem. When we see that a car is moving, we can hear the engine and road sounds at the same time. 
The goal of our work is to learn to recover the coordinates of moving vehicles purely from stereo sound, without any visual input. Such a system has a variety of practical applications.
For instance, a traffic monitoring system could be deployed using just microphones, which are less expensive, lower power, privacy-preserving and require less bandwidth than cameras (the camera would only be required during a brief training phase). Likewise, co-registered audio-visual localization could be used to augment visual tracking in a robot, allowing it to perform well even under the poor lighting conditions.

Directly training an audio-only localization system in a supervised setting is cumbersome, since manually associating bounding boxes of objects with their corresponding audio would require extensive, labor-intensive manual annotation. Instead, we capitalize on the natural correspondence of audio and visual streams contained in the unlabeled videos, using a self-supervised training approach. Intuitively, our system can learn to localize moving objects by seeing and hearing the object move simultaneously. Our proposed framework, which we refer to as \myapproach, is built on a student-teacher training procedure~\cite{aytar2016soundnet,owens2016ambient,albanie2018emotion,gan2018geometry,gupta2016cross}. It consists of a vision teacher network and a stereo-sound student network, enabling object detection knowledge to be transferred across modalities during training time. Specifically, we first use the vision teacher network to detect the objects (in this case, moving vehicles) in videos, and we train a stereo-sound network that maps the audio signals to the bounding box coordinates predicted by the vision network.  
Then at test time, the student sound network can directly predict object coordinates from sounds. We evaluate our \myapproach approach on a newly collected \mydataset dataset. Our results show that the proposed system significantly outperforms several baseline approaches, measured by a set of existing metrics in computer vision. In summary, our work makes the following contributions:

\begin{compactitem}
	\item To the best of our knowledge, we are the first to approach the  problem of localizing objects in a visual reference frame, purely from audio signals.

	\item We propose to leverage the correspondences between vision and sound in the unlabeled videos as supervision to train a network that can transfer the knowledge of object locations from the visual modality to the sound modality.
	
	\item  We have collected and annotated a new \mydataset dataset for this new task. We expect that this dataset can help advance research in the area of cross-modal (vision+audio) perception.

    \item We demonstrate that the proposed \myapproach system works well for localizing vehicles through sound alone, and even outperforms direct visual tracking under poor lighting conditions. 
\end{compactitem}

\section{Related Work}
\label{sec:related}

Our work can be uniquely positioned in the context of two recent research directions: sound localization and cross-modal learning.

\subsection{Sound Localization}

Localization using sound is a well-established area of study.  Some organisms and man-made systems use active techniques for sound localization and auditory scene perception. Echolocation involves emitting sound waves and analyzing the returning reflected sound waves to estimate the distances of obstacles. Echolocation is commonly observed in animals that operate in dark or turbid environments, \textit{e.g.} bats and dolphins rely on echolocation to position themselves and to locate prey. Based on the same principles, engineers have designed sonar (Sound Navigation and Ranging) systems ~\cite{winder1975ii}. Sonar is especially common in underwater and robotics applications ~\cite{leonard2012directed,teng2017hearing}.

Passive audio localization technology typically involves using microphone arrays and beam-forming techniques~\cite{brandstein2013microphone}. The timing differences in the sound received by the different microphones can be used to estimate the location of the sound. Even smaller devices such as smart home speakers often use several microphones in order to improve sound quality. For example, \cite{sainath2017multichannel} developed techniques to improve automatic speech recognition accuracy using multichannel audio inputs. Using multichannel audio has also been shown to be advantageous in other scenarios, such as sound source separation \cite{nugraha2016multichannel}. Our work here uses a stereo microphone system, which is the simplest system that can take advantage of spatial measurement of sound for localization. 

The present work is also related to previous work in localizing sounds in visual inputs~\cite{hershey2000audio,fisher2001learning,izadinia2013multimodal,barzelay2007harmony,ban2018variational,li2018online,arandjelovic2017objects,senocak2018learning}, which aims to identify which pixels in a video are associated with an object making a particular sound. 

Recent approaches~\cite{arandjelovic2017objects,senocak2018learning,zhao2018sound,gao20182} have trained a deep neural network to measure the correlations between visual and sound, and then used network localization techniques to locate the sound source in images.  Where this past work sought to localize sound sources in images when both visual and audio inputs are present, here we instead seek to locate objects within a visual reference frame using audio inputs only at test time.

\subsection{Cross-modal Self-supervised Learning}

Our work is in the domain of self-supervised learning, which exploits implicit labels that are freely available in the structure of the data. Audio-visual data offers a wealth of resources for knowledge transfer between different modalities~\cite{aytar2016soundnet,castrejon2016learning,salvador2017learning,socher2013zero}. Our work is also closely related to the student-teacher learning paradigm~\cite{ba2014deep,hinton2015distilling,aytar2016soundnet,DBLP:conf/mm/AlbanieNVZ18,gan2016you}, where a student network attempts to mimic the teacher network outputs. For example, \cite{owens2016ambient} used sound signals as supervision to train visual networks, \cite{aytar2016soundnet} used visual features to supervise the learning of audio networks, and \cite{arandjelovic2017look,owens2018audio} used sound and vision to jointly supervise each other. \cite{gao20182, morgado2018self} also explored how to generate spatial sound for videos. More recently, \cite{zhao2018sound,zhao2019sound,ephrat2018looking,gao2018learning,rouditchenko2019self} used the visual-audio correspondence to separate sound sources. In contrast to previous work that has only transferred class-level information between modalities, this work transfers richer, region-level location information about objects.

\begin{figure*}[!ht]
	\centering
	\includegraphics[width = 1\linewidth]{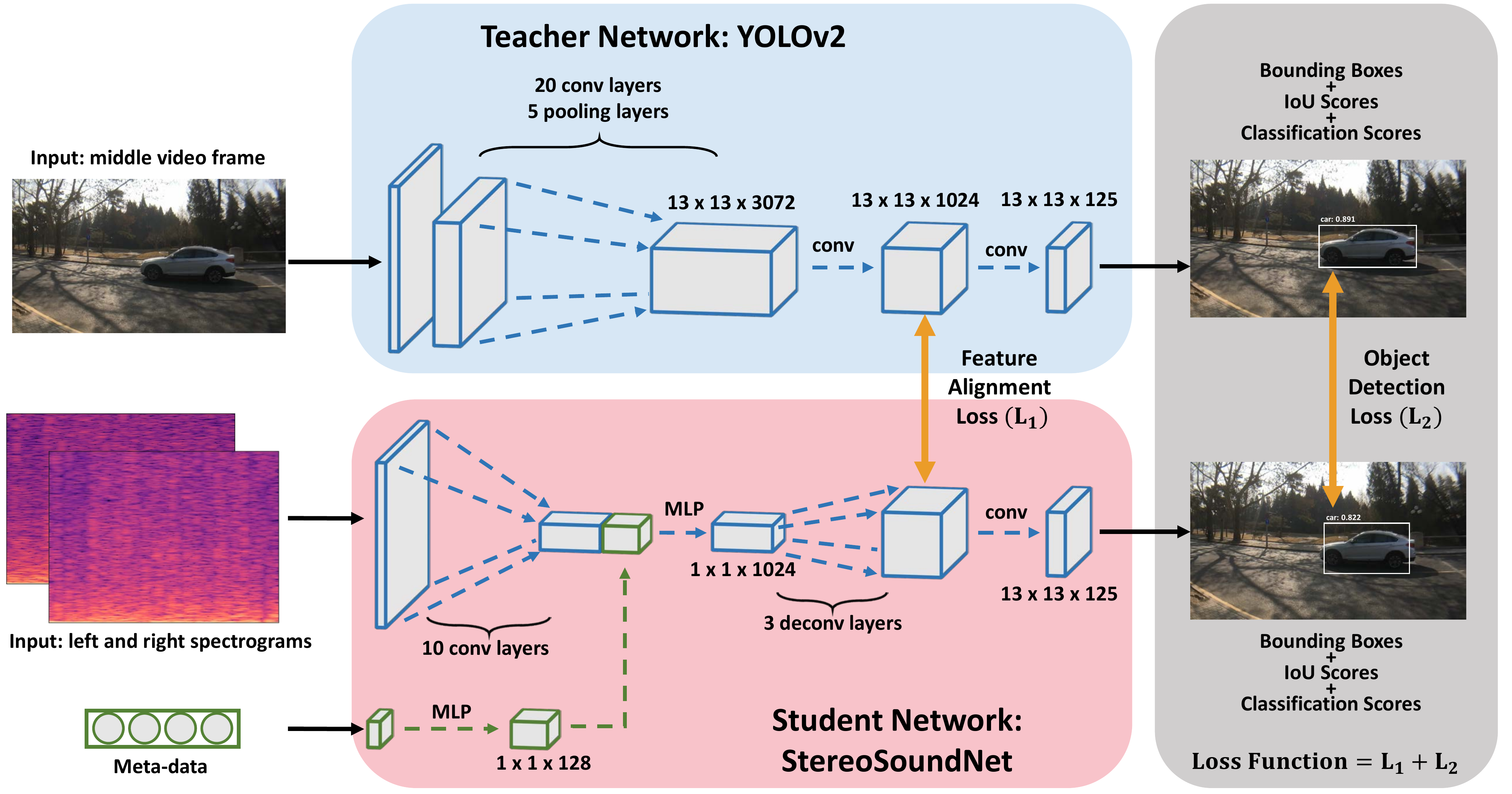}
	
	\caption{The framework of \myapproach. We illustrate the learning phase by jointly using sounds and frames from videos. We first decompose a video into several video segments, where each is 1 seconds long. During training, a pre-trained YOLOv2 network predicts bounding boxes of the middle video frame as pseudo-labels, while the auditory student network takes pre-computed spectrograms of sounds and camera meta-data as input to regress that pseudo-labels and also to align the internal feature representation of the vision network. During testing, the auditory network can work independently to detect vehicles.}
	\label{fig:stereonet}
	
\end{figure*}

\section{Approach}
\label{sec:approach}
Central to our approach is the observation that the natural synchronization between vision and sound in unlabeled video can serve as a form of self-supervision for learning. A machine could therefore learn to predict the location of an object by seeing and hearing many examples of moving vehicles that produce sound. We model the learning problem using a student-teacher framework. Our system is trained simultaneously using video frames and sounds, which allows the auditory student network to learn how to localize vehicle bounding boxes from the visual teacher network.

We first introduce the building blocks of our \myapproach system, and then we present how to transfer the knowledge in the visual vehicle detection model to the sound signals given  the camera meta-data by training the audio subnetwork using object detection loss and feature alignment constraint. Finally, we present a temporal smoothing approach to track the vehicles over the time. We outline the framework of our proposed approach in Figure~\ref{fig:stereonet}.

\subsection{Network Architectures}
Our auditory object localization system is composed of two key components: a teacher vision subnetwork and a student audio subnetwork. 

\noindent\textbf{Vision Subnetwork.}
We adopt the YOLOv2~\cite{redmon2017yolo9000} for the vision-based teacher network, since it offers a good trade-off between the speed and accuracy for object detection. YOLOv2~\cite{redmon2017yolo9000} is a modification of YOLO~\cite{redmon2016you} with batch normalization, high image resolution, convolutional with anchor boxes and multi-scale training, which is thus capable to simultaneously predict multiple bounding boxes and their class probabilities directly from full images in a single stage. 

The backbone of YOLOv2 is a Darknet, which consists of 19 convolutional layers and 5 max pooling layers. To make it more suitable for the object detection, the last convolutional layer is replaced by three 3$\times$3 convolutional layers with 1024 filters, followed by a 1$\times$1 convolutional layer with the number of outputs to be detected.  Similar to the identity mappings used in ResNet, there is also a pass-through layer from the final 3$\times$3$\times$512 layer to the second to last convolutional layer to aggregate the fine-grained level features.  To make the model more stable and easier to learn, the network is trained to predict the location coordinates relative to the location of the anchor boxes.

To prepare the data, we first decompose each video clip into several $T = 1s$ video segments\footnote{ The localization results could be improved with the increasing length of the video segments, but the performances remain stable for the segment longer than 1 second.}, and then pick the middle frame of each segment as the input to the teacher network. During training, each middle video frame is fed into a YOLOv2 model pre-trained on Pascal VOC 2007 and VOC 2012 dataset to obtain the vehicle detection results. In order to make the detection results smoother, we also apply non-maximum suppression (NMS) as the post-processing. 

\noindent\textbf{Audio Subnetwork.}
We cast object detection from the stereo sound as a regression problem. We take the object detection results produced by the teacher vision subnetwork as a pseudo-labels, and then train a student audio subnetwork (StereoSoundNet) to regress the pseudo bounding box coordinates directly from the audio signals. Considering different camera angle might bring relatively larger change to visual content than the audio, we resolve it by explicitly taking the meta-data of the camera as input when training the StereoSoundNet. The meta-data here includes the camera height, pitch angle, and orientation between the camera and a street.  

We first convert each 1-second audio segment into spectrograms through Short-Time Fourier Transform (STFT). Since there are two channels in the stereo sounds, we compute their spectrograms separately and then stack them as the input to the StereoSoundNet. To transform the F-T (frequency-time) representations of the input audio spectrogram to the view of the camera, we first use 10 strided convolutional layers, where each is followed by a batch normalization layer and a ReLU activation function, as an encoder to compress the stereo sound signals as a 1$\times$1$\times$1024 feature map, removing the spatial resolution. And then we employ a multilayer perceptron to encode the meta-data into a 1$\times$1$\times$128 feature map.  After concatenating the compressed sound information and encoded meta-data channel-wise, a decoder, which consists of 2 fully connected layers and 3 de-convolutional layers, is used to reconstruct the spatial resolution and map the audio information to the camera view. The final output is similar to the YOLOv2 and we adopt the object detection loss used in YOLOv2 to train the StereoSoundNet.

\subsection{Transfer of Knowledge from Vision to Sound}
In order to transfer knowledge from vision object detection models into the sound modality, we use both the object detection loss and feature alignment loss to train the StereoSoundNet.

\noindent\textbf{Transfer object detection classifiers.} During training, we take the output of a well-established vision-based YOLOv2 object detection model, and then train the audio subnetwork to recognize and localize the objects.
Concretely, we train the audio subnetwork using three loss constraints as suggested in the~\cite{redmon2017yolo9000}, including bounding box IoU prediction, bounding box coordinate regression, and class probabilities prediction. In our case, we predict 5 boxes for each location on the output feature map, with 4 coordinates, 1 Intersection over Union (IoU), and 20 class probabilities for each box. So the output of the audio subnetwork is of size H$\times$W$\times$125.

\noindent\textbf{Alignment of Feature Representations.} We additionally add a feature representation alignment constraint into the training loss. The observations in~\cite{aytar2017see} indicate that the internal high-level representations for the emerging objects can be shared across modalities, even though each input has its own distinct features in the early stage of the network. We expect the feature representations of two modalities to be close enough under certain distance metrics. 

Following~\cite{aytar2017see}, we use the ranking loss to constrain the features. Specifically, the feature alignment loss is

\begin{align}
    \sum_i^N \sum_{j \ne i} \max\{0, \Delta - \psi(f_{s_i}, f_{v_i}) + \psi(f_{s_i}, f_{v_j})\},
    \label{eqn:ranking}
\end{align}
where $N$ is the number of training samples in one mini-batch, $\Delta$ is a margin hyper-parameter, $\psi$ is a similarity function, and $j$ iterates over negative examples in the mini-batch. Here, $f_{s_i}$ and $f_{v_i}$ indicate the predicted feature representation of the $i^{th}$ sound clip from the student audio subnetwork and the corresponding feature representations from the teacher vision subnetwork respectively. This loss function encourages both aligned features for the paired input and the discriminative features for the unpaired one.
As for similarity function $\psi$, we choose L-2 distance. That is 
\begin{equation}
	\psi(a, b) = ||a - b||_2.
\end{equation}

\noindent\textbf{Training and Inference.}
When training the audio subnetwork, the goal is to enforce both the internal feature representations  and the final bounding box predictions of the audio student subnetwork as close as the vision teacher subnetwork. We do not update vision subnetwork in the training phase. During testing, the audio subnetwork can work independently, straight from the sound to bounding box locations and class probabilities. We keep the boxes whose confident scores are higher than 0.5 as predicted bounding boxes. If all confident scores are lower than 0.5, we select one box with the highest score. Similar to vision subnetwork, NMS is applied as the post-processing to eliminate the repeated detected boxes.

\noindent\textbf{Tracking.}
In order to create tracklets of vehicles in videos, we proposed a tracking by IoU approach to aggregate the objects bounding boxes across time. Specifically, we keep the top 5 object proposal based on the confidence scores of the StereoSoundNet on each frame. We then initialize a tube if any proposal box's confidence score is higher than a threshold $\tau_1$ (we set $\tau_1 = 0.7$).  For each frame, it could be possible to have more than one tube.  We decide the next bounding box in each tube by calculating the IoU score between that box and 5 proposal boxes in the next frame, and then select the box with the highest confident score in the next frame if its IoU score is higher than a threshold $\tau_2$ (we set $\tau_2 = 0.4$). We then use the selected bounding boxes to update that tube. If no boxes are selected, we end that tube. We only save the tube if the confidence scores of first two boxes are both higher than a threshold $\tau_3$ (we set $\tau_3 = 0.4$). This strategy can remove some incorrectly initialized tubes. Finally, we apply an exponential smoothing over all the frames in videos to obtain the tracklets in videos.

\begin{table*}[t]
\begin{center}
\begin{tabular}{l||c|c|c|c|c}
Approach             & AP@Ave         & AP@0.5         & AP@0.75       & $CD_x$          & $CD_y$         \\ \hline\hline
YOLOv2        & 42.39            & 79.54            & 41.62        & 6.46\%      & 2.55\%             \\ \hline \hline
Random                & 0.00           & 0.00           & 0.00          & 33.44\%          & 26.09\%         \\ 
Waveform              & 5.87           & 23.14           & 0.91          & 15.63\%          & 5.18\%          \\ 
Mono                  & 11.80           & 38.57           & 3.31          & 14.49\%          & 4.68\%          \\ \hline \hline
w/o feature alignment & 21.55          & 57.47          & 10.01          & 10.82\%          & 4.06\%          \\  
w/o meta-data & 9.45          & 27.76         & 3.43          & 13.79\%          & 12.26\%          \\ \hline \hline
Ours                  & 21.55 & 57.47 & 10.13 & 10.53\%          & 3.98\% \\ 
Ours (w Tracking)      & \textbf{25.05}        & \textbf{60.70}   &  \textbf{15.96}       &  \textbf{7.76\%}  & \textbf{3.75\%} \\ \hline

\end{tabular}
\end{center}
\caption{Compared results of \myapproach in term of Average Precision (AP) and Center Distances (CD). \textbf{Higher} AP number indicates \textbf{Better} results. \textbf{Lower} Center Distances (CD) number indicates \textbf{Better} results.}
\label{tab:main}
\end{table*}

\section{Experiments}
\label{sec:experiment}
In this section, we first present a newly collected \mydataset dataset, and then evaluate the performance of our proposed \myapproach on it. We contrast our algorithm to several competing baselines. We also demonstrate that our \myapproach is more robust than the visual tracking system under the poor lighting condition. Finally, we examine the cross-scene generalization abilities and visualize some  \myapproach results.

\subsection{Dataset}
\begin{figure}[!t]
	\centering
	\includegraphics[width = 1\linewidth]{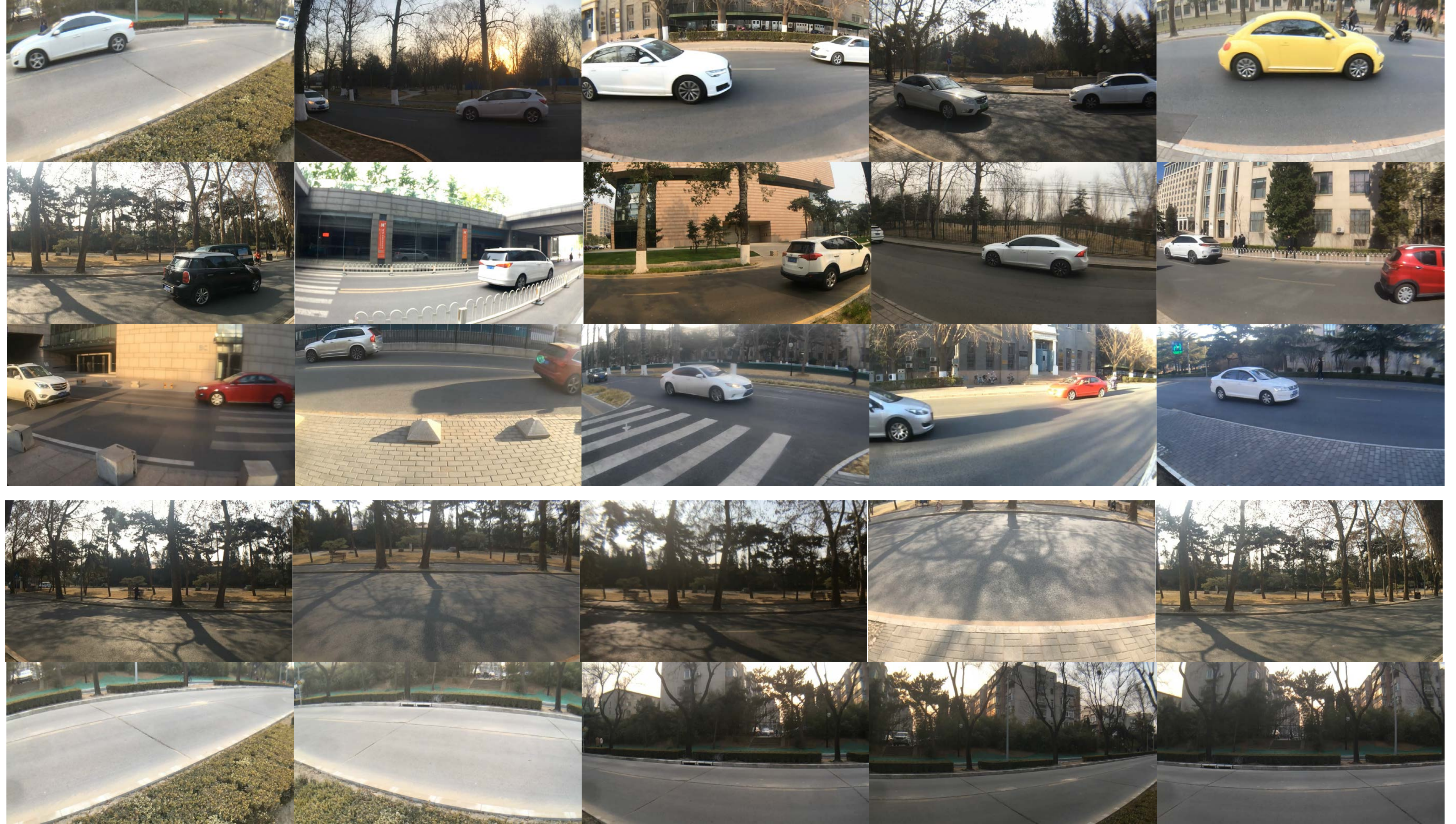}
	\caption{Examples of the scenes in our dataset.}
	\label{fig:datasets}
\end{figure}

We collected a new \mydataset dataset using a portable setup, composed of a smartphone and a Shure MV88 digital stereo condenser microphone to record the stereo sound. We attached a wide angle lens to the smartphone to increase the field of view. Videos were recorded on 15 different streets. We also adjusted the camera's height, pitch angle, and orientation of the camera relative to the road to capture more diverse videos. The height of the camera was varied from 0 to 2 meters with the pitch angle and the rotate angle in the range of [$-30^{\circ}$, $+30^{\circ}$] and [$-35^{\circ}$, $+35^{\circ}$]\footnote{The negative sign indicates the downward direction in pitch angle and the left direction in rotate angle.}, respectively. For each street, we randomly selected 6 camera angles within the range mentioned above. In the first three rows of Figure~\ref{fig:datasets}, we present 15 different scenes in our dataset. As for the last two rows, each of them shows 5 randomly chosen angles of the same scenes.

During video capture, we avoided parked vehicles that do not make any sound in the scene. We also excluded video clips that contained more than two vehicles, since multiple car detection is still challenging. Therefore, after dataset post-processing, the raw videos are cropped into 3,243 short video clips (around 3 hours in total), which contain two cases: single car and two car conditions. The audio was recorded at a sampling rate of 48kHz in stereo format.

\subsection{Experimental Setup}

\noindent\textbf{Data Split.}
We split the video clips into three parts: 3,329 for training, 415 for validation, and 423 for testing. We further decompose video clips into a number of 1-second video segments as training and testing samples, leading to 227,810 samples for training, 27,779 samples for validation and 28,672 samples for testing. 

To collect an unbiased testing set for the evaluation, we label the ground-truth bounding boxes location of the vehicle on the middle frames of each testing sample using Amazon Mechanical Turk (AMT). To be noted, the manually labeled testing data are only used for the evaluation, not for any model training.

\noindent\textbf{Evaluation Metric.}
To evaluate our method, we used the traditional object detection evaluation metric, Average Precision (AP). We report the AP at IoU 0.5 and 0.75 and average AP across IoU thresholds from 0.5 to 0.95 with an interval of 0.05.  
We used the center point of the predicted box to measure the localization accuracy on $x$ and $y$ coordinates. Specifically, supposing $(P^x,P^y)$ is the center point of the closest predicted box to the ground-truth box $(G^x,G^y)$, we define the Center Distances (CD) of $x$ and $y$ coordinates as $CD_x = \frac{1}{K} \sum_i^K |P^x_i-G^x_i|/w$ and $CD_y = \frac{1}{K} \sum_i^K |P^y_i-G^y_i|/h$, where K is the total number of ground-truth box and $w$ and $h$ are the width and height of the video.

\noindent\textbf{Implementation Details.} We train the StereoSoundNet for 60 epochs with initial learning rate as 0.0001, dividing it by 10 every 20 epochs. The batch size is set to 80 and we use the stochastic gradient descent optimization with a weight decay of 0.0005 and momentum of 0.9. The feature alignment loss is implemented on the last feature map of the teacher and student network and the margin hyper-parameter $\Delta$ is set to 0.2. The total loss is the sum of feature alignment loss and object detection loss, which have the same weight during training.

All videos in the datasets are in 24 fps with 1280$\times$720 resolution. Each training or testing sample is a 1-second video segment containing 24 frames and 1-second stereo sound. To generate spectrogram, we first normalize the raw waveform with its maximum value, then an STFT with a window size of 1024 and a hop length of 256 is computed on the normalized waveform. We further re-sample it on a Mel-Frequency scale with 80 frequency bins, resulting in a 187$\times$80 Time-Frequency (T-F) representation of the sound. We re-size spectrogram and the RGB frame to 256$\times$256 and 416$\times$416, respectively. The meta-data is normalized to [0,1] for stable training.

\subsection{Baselines}

To evaluate our framework, we compare against alternative approaches as baselines:
\begin{compactitem}
\item \textbf{Random:}  We randomly draw 1 or 2 boxes with random size within the frame as auditory vehicle detection results.

\item \textbf{Mono sound:} For each audio clip, we simply add the two channels into one channel, convert them into one spectrogram and then feed them to the audio subnetwork. We maintain other parts same as the StereoSoundNet.

\item \textbf{Raw Waveform:} We apply the SoundNet~\cite{aytar2016soundnet} architecture with the raw stereo sound waveform as the input, instead of extracting the spectrogram from the audio. We train the SoundNet using both the object detection loss and feature alignment loss as well.

\item  \textbf{W/O Meta-Data:} 
We use the same encoder-decoder based framework. We first use a 10-layer CNN to encode the spectrogram into a vector and then use a de-convolution network to map the vector to the object bounding boxes. Similar to the StereoSoundNet, we also use both the object detection loss and feature alignment loss to train the audio subnetwork.

\item  \textbf{W/O Feature alignment:}  We exclude the feature alignment loss during training of the audio subnetwork. Thus the audio stream directly learns to regress the bounding boxes generated by the teacher vision subnetwork.

\end{compactitem}

\subsection{Experimental Results}

\subsubsection{Results Analysis}

Comparisons of results with baseline approaches are reported in Table~\ref{tab:main}. Unless otherwise specified, the reported results do not consider tracking post-processing. It is clear from the table that when our \myapproach is trained with both the object detection loss and the feature alignment constraint, it outperforms all the audio only baselines.  Using tracking post-processing further increases the performance of AP and also leads to more consistent and smooth tracking. The oracle vision-based YOLOv2 achieved 79.54\% in terms of AP@0.5. Our StereoSoundNet still has around 20\% performance gap. We think more training data and better microphone might further reduce the gap. We leave these to future work.

\begin{table}
\begin{center}
\begin{tabular}{c|c|c}
The Number of vehicles             &  Single       &  Multiple        \\ 
\hline\hline
AP@Ave  & 26.53 & 11.58       \\ \hline
AP@0.5 & 70.12 &  32.06      \\ \hline
AP@0.75 & 12.61 & 5.21 \\ \hline

\end{tabular}
\end{center}
\caption{Auditory vehicle localization results of single and multiple vehicles in terms of Average Precision (AP) and Center Distances (CD).  \textbf{Higher} AP number indicates \textbf{better} results. \textbf{Lower} Center Distances (CD) number indicates \textbf{better} results.}
\label{tab:multiple}
\end{table}

We also report the results on single vehicle and multiple vehicles cases respectively. The results are shown in Table~\ref{tab:multiple}. Although performance drops on multiple vehicles cases, it is still able to produce convincing localization prediction. These results indicate \myapproach can implicitly perform the sound separation and then localize different moving vehicles simultaneously.

\subsubsection{Spectrogram \textit{v.s.} Raw Waveform}
We also compared two sound representations (\ie spectrogram and raw waveform) that are commonly used in the context of cross-modal learning.  We use a SoundNet~\cite{aytar2016soundnet} pre-trained on the large-scale audio dataset  as the base model of the audio subnetwork and then fine-tune it on our car tracking data. 

Results in Table~\ref{tab:main} show that the spectrogram sound representations clearly outperform the raw waveform format.  We speculate that the spectrogram tends to more directly capture frequency differences contained in audio, which is critical for the vehicle localization using sound only.

\begin{figure}
	\centering
	\includegraphics[width = 1\linewidth]{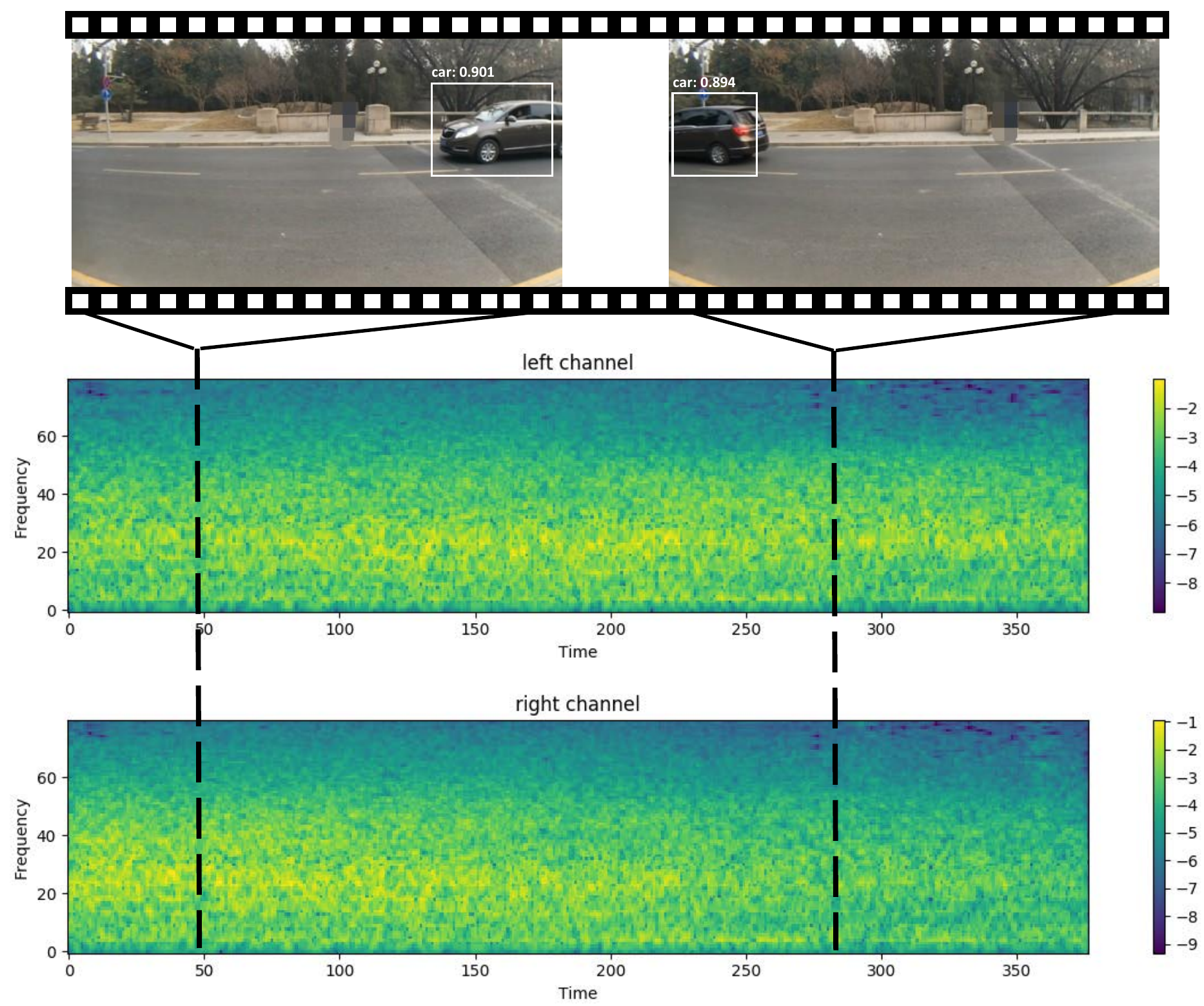}
	
	\caption{Visualization of \myapproach results of one video clip and its corresponding input spectrograms.}
	\label{fig:spectrogram}
\end{figure}

\subsubsection{Mono Sound \textit{v.s.} Stereo Sound}
We further examine whether the stereo sound is necessary for the \myapproach to learn how to localize targets. Specifically, we compare, in Table~\ref{tab:main}, a  baseline training the audio subnetwork with the mono sound. We simply add the two channels of stereo sound and then convert it to a spectrogram.

We observe that the results are significantly worse with mono sound as compared to stereo sound in terms of AP score.  We also observe that it is impossible to predict the vehicle coming from the left or right just based on the sound volume change over time. This indicates that stereo sound provides stronger supervision for localization.  In Figure~\ref{fig:spectrogram}, we also visualize the input spectrograms and the corresponding stereo sound localization results. At the beginning of the video, there is a car on the right side of the frame, and it is clearly observed that the amplitude of the right channel's spectrogram is higher than the left channel. As the car moves to the left side, the amplitude of the right channel goes down while the opposite trend is

\begin{figure*}[!ht]
	\centering
	\includegraphics[width = 0.95\linewidth]{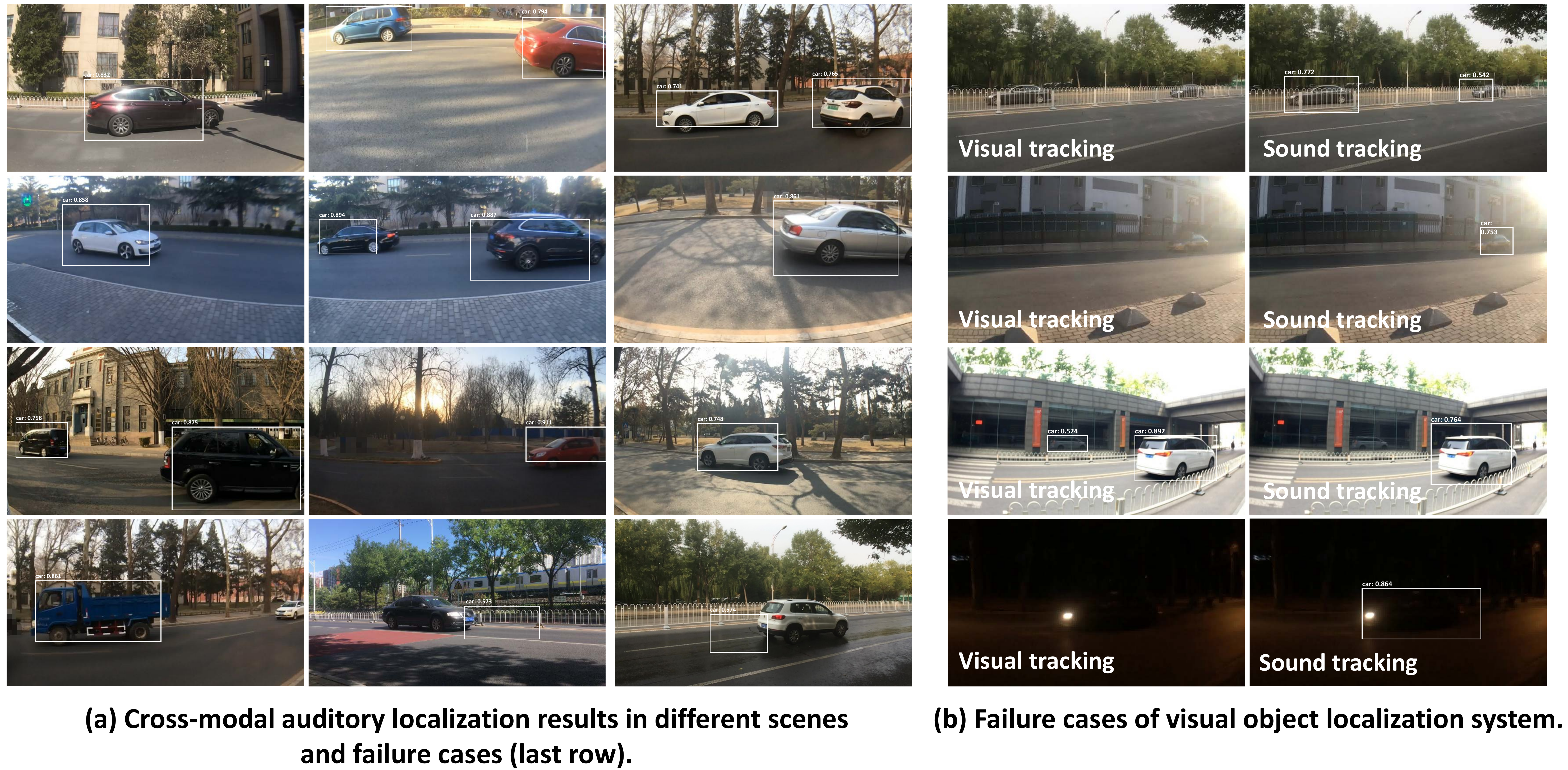}
	
	\caption{(a) Visualization of \myapproach results in different scenes together with the failure cases due to the noisy sounds from trucks, train and sediment. (b) The common failure examples using visual object localization system.}
	\label{fig:visualization}
\end{figure*}

\begin{table}[h]
\begin{center}
\scalebox{0.8}{
\begin{tabular}{l||c|c|c|c|c}
Approach         & MOTA $\uparrow$    & ID Sw. $\downarrow$ & Frag.$\downarrow$ & FP $\downarrow$ & FN $\downarrow$  \\ \hline \hline

Ours       & 13.9\%          & 1100              & 1318             & 9974         & 13462          \\ \hline
Ours (w Tracking)     & \textbf{18.7\%} & \textbf{954}              & \textbf{502}             & \textbf{9016}         & \textbf{12868} \\ \hline

\end{tabular}}
\end{center}
\caption{Compared results in term of tracking metrics.  ``$\uparrow$'' means that higher is better and ``$\downarrow$'' represents that lower is better.}
\label{tab:tracking}
\end{table}

\subsubsection{Tracking Performances}
 In order to measure the tracking performance, we leverage multiple object tracking accuracy (MOTA), identity switches (ID Sw.), fragment (Frag.), false positive (FP) and false negative (FN) as evaluation metrics~\cite{bernardin2008evaluating,5206735}. For the baseline without tracking post-processing, we randomly assign an ID to each box as such baseline cannot predict ID. Results are shown in Table \ref{tab:tracking}. 
Using tacking post-processing achieves better MOTA and ID Sw., which shows its superiority in detecting objects and keeping their trajectories. It is worth noting that random ID assignment is a strong baseline as it does not cause ID switches under the single car condition.
The better Frag., FP, and FN, which are independent to ID assignment, indicates that with tracking post-processing, our model incurs fewer switches from tracked to not tracked, less false positive detection and less missing objects.

\begin{table}
\begin{center}
\begin{tabular}{l||c}
Approach                     & AP@0.5                   \\ \hline\hline
Random                                    & 0.00               \\ \hline
Yolov2 (vision)                        & 6.78              \\ \hline
Ours  (audio)        & \textbf{30.88}  \\ \hline
\end{tabular}
\end{center}
\caption{Auditory vehicle localization results under poor lighting conditions in terms of Average Precision (AP). \textbf{Higher} AP number indicates \textbf{better} results}
\label{tab:poor}
\end{table}
\subsection{Performances Under the Poor Lighting}
We conduct additional experiments to evaluate whether our auditory object tracking is still robust under poor lighting conditions, where tradition vision-based object tracking typically fails. We first collect 5 videos at night and then label the object localization on the key frames using Amazon Mechanical Turk (AMT) for the evaluation. 

 We directly applied the StereoSoundNet trained on daytime data to the nighttime scenario without any fine-tuning. The results are reported in Table~\ref{tab:poor}. It is not surprising that the visual tracking system fails in these scenarios, due to the fact that vision-based algorithms are very sensitive to poor lighting. However, we observe that our \myapproach maintains robust tracking performance, compared with the vision-based system. We also visualize some intriguing examples in Figure~\ref{fig:visualization}.  More tracking results can be viewed in demo videos.

\begin{table}
\begin{center}
\begin{tabular}{c|c|c}
              &  w meta-data       &  w/o meta-data        \\ 
\hline\hline
AP@Ave  & 12.24 & 0.00       \\ \hline
AP@0.5 & 42.79 &  0.00      \\ \hline
$CD_x$  & 10.17\% & 43.53\% \\ \hline
$CD_y$ & 5.02\% & 35.23\% \\ \hline
\end{tabular}
\end{center}
\vspace{-7pt}

\caption{Generalization of auditory vehicles detection system. \textbf{Higher} AP number indicates \textbf{better} results. \textbf{Lower} Center Distances (CD) number indicates \textbf{better} results.}
\vspace{-15pt}
\label{tab:generlization}

\end{table}

\subsection{Generalization on Novel Scenes}
One benefit of using camera meta-data for auditory vehicle localization is that it allows better generalization to novel scenes, as meta-data can explicitly provide the camera's position when the visual reference frames were captured. In Table \ref{tab:generlization}, we explore the generalization of our auditory object detection system  by comparing performances on new scenes. Specifically, we split the videos collected in 15 scenes into two disjoint parts: 10 scenes for training and the other 5 scenes for testing. Note that the camera shots of the testing data may be different from the training data. We observe that with the help of prior knowledge about the camera height and angle, generalization could be significantly improved, but still have considerable performance gaps with seen scenarios.

\subsection{Visualization}

We visualize some \myapproach results under different scenes with different camera positions on Figure~\ref{fig:visualization} (a). The Figure demonstrates that our StereoSoundNet performs robustly in different scenes with different camera angles using only stereo sound and camera meta-data as input. We also observe some failure cases on fast moving vehicles and noisy sound (\eg. construction, wind and sediment).  Figure~\ref{fig:visualization} (b) reveals some interesting cases we found in our datasets that visual object localization framework fails to track the moving cars due to occlusion, backlighting, reflection, and bad lighting condition, while our StereoSoundNet succeeds. Our proposed \myapproach system has good potentials to assist in the visual localization of objects in these cases of less-than-ideal image quality.

\section{Conclusion}
\label{sec:conclusion}

In this work, we leverage stereo sound to perform \myapproach. We created a new \mydataset dataset that consists of over 3000 video clips for studying this task. We also provide an automatic quantitative method to evaluate the models and the results. To address this challenging problem, a novel student-teacher based network is proposed, which can successfully transfer knowledge from a vision-based object detection network to the sound modality.  The new auditory vehicle tracking algorithm also demonstrates its potentials to augment visual tracking systems under poor light conditions.  Areas for future work include extending our approach to more simultaneous scenes and to more different kinds of objects.

{\small
\bibliographystyle{ieee}
\bibliography{ref}
}

\end{document}